%% file: main.tex
\documentclass[10pt,twocolumn,letterpaper]{article}

\usepackage{cvpr}
\usepackage{times}
\usepackage{epsfig}
\usepackage{graphicx}
\usepackage{amsmath}
\usepackage{amssymb}

\usepackage{verbatim}
\usepackage{helvet}
\usepackage{courier}
\usepackage{amsfonts}
\usepackage{subfigure}
\graphicspath{ {./images/} }
\usepackage{booktabs}
\usepackage{threeparttable}
\usepackage{multicol}
\usepackage{multirow}
\usepackage{tabularx}
\usepackage[bottom]{footmisc}
\usepackage{lipsum}
\newcolumntype{L}[1]{>{\raggedright\arraybackslash}p{#1}}
\newcolumntype{C}[1]{>{\centering\arraybackslash}p{#1}}
\newcolumntype{R}[1]{>{\raggedleft\arraybackslash}p{#1}}

\DeclareMathAlphabet\mathbfcal{OMS}{cmsy}{}{n}

\usepackage[pagebackref=true,breaklinks=true,letterpaper=true,colorlinks,bookmarks=false]{hyperref}

\cvprfinalcopy 

\def\cvprPaperID{5118} 

\ifcvprfinal\fi
\begin{document}

\title{PVN3D: A Deep Point-wise 3D Keypoints Voting Network for 6DoF Pose Estimation}

\author{Yisheng He$^{1}$ \quad {Wei Sun}$^{2}$ \quad {Haibin Huang}$^{3}$ \quad {Jianran Liu}$^{2}$ \quad {Haoqiang Fan}$^{2}$ \quad {Jian Sun}$^{2}$ \\
	${^1}$Hong Kong University of Science and Technology \\
	${^2}$Megvii Inc.\quad \quad
	${^3}$Kuaishou Technology \\
}

\maketitle

\thispagestyle{empty}
\pagestyle{empty}

\input{abs}
\input{intro}
\input{related_work}

\input{method}

\input{results}

\input{discussion}
\appendix
\input{appendix}

{\small
\bibliographystyle{ieee}
\bibliography{ref}
}

\end{document}

%% file: abs.tex
\begin{abstract}
In this work, we present a novel data-driven method for robust 6DoF object pose estimation from a single RGBD image. Unlike previous methods that directly regressing pose parameters, we tackle this challenging task with a keypoint-based approach. Specifically, we propose a deep Hough voting network to detect 3D keypoints of objects and then estimate the 6D pose parameters within a least-squares fitting manner. Our method is a natural extension of 2D-keypoint approaches that successfully work on RGB based 6DoF estimation. It allows us to fully utilize the geometric constraint of rigid objects with the extra depth information and is easy for a network to learn and optimize. Extensive experiments were conducted to demonstrate the effectiveness of 3D-keypoint detection in the 6D pose estimation task. Experimental results also show our method outperforms the state-of-the-art methods by large margins on several benchmarks. Code and video are available at \url{https://github.com/ethnhe/PVN3D.git}.


\end{abstract}

%% file: intro.tex
\section{INTRODUCTION}

In this paper, we study the problem of 6DoF pose estimation, i.e. recognize the 3D location and orientation of an object in a canonical frame.  It is an important component in many real-world applications, such as robotic grasping and manipulation \cite{collet2011moped,tremblay2018deep,zhu2014single}, autonomous driving \cite{geiger2012we,chen2017multi,xu2018pointfusion}, augmented reality \cite{marchand2015pose} and so on.   

\newcommand\blfootnote[1]{%
  \begingroup
  \renewcommand\thefootnote{}\footnote{#1}%
  \addtocounter{footnote}{-1}%
  \endgroup
}
\blfootnote{This work is supported by The National Key Research and Development Program of China (2018YFC0831700).}

\begin{figure}
    \centering
    \includegraphics[scale=0.4]{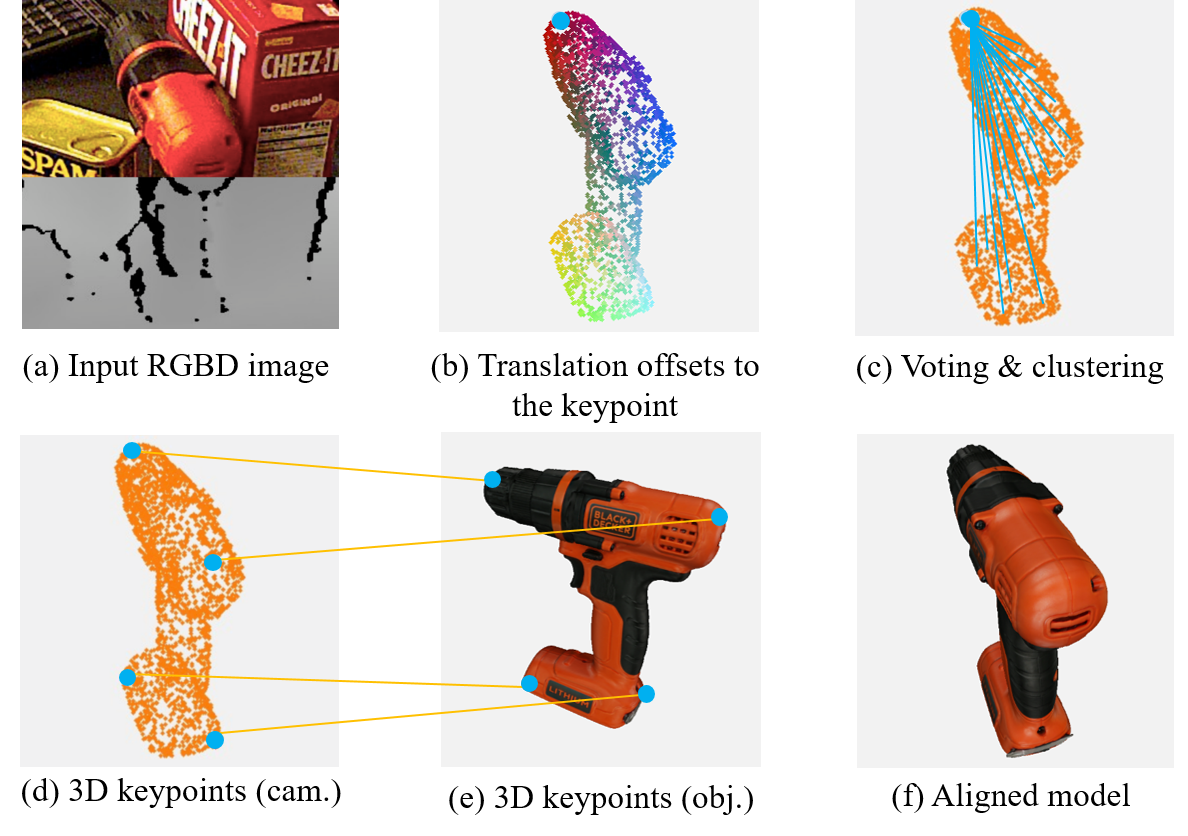}
    \caption{
        \textbf{Pipeline of PVN3D:}  With an input RGBD image (a), we use a deep Hough voting network to predict the per-point translation offset to the selected keypoint (b). Each point on the same object votes for the selected keypoint and the center of the cluster is selected as a predicted keypoint (c). A least-squares fitting method is then applied to estimate 6D pose parameters (d)-(e). The model transformed by estimated pose parameters is shown in Figure (f).
    }
    \label{fig:our_approach}
\end{figure}

6DoF estimation has been proven a quite challenging problem due to variations of lighting, sensor noise, occlusion of scenes and truncation of objects. Traditional methods like \cite{hinterstoisser2012model,lowe1999object} used hand-crafted features to extract the correspondence between images and object mesh models. Such empirical human-designed features would suffer from limited performance with changing illumination conditions and scenes with heavy occlusion. More recently, with the explosive growth of machine learning and deep learning techniques, Deep Neural Network (DNN) based methods have been introduced into this task and reveal promising improvements. \cite{wang2019densefusion,xiang2017posecnn} proposed to regress rotation and translation of objects directly with DNNs. However, these methods usually had poor generalization due to the non-linearity of the rotation space explained by \cite{peng2019pvnet}.
Instead, recent works utilized DNNs to detect 2D keypoints of an object, and computed 6D pose parameters with Perspective-n-Point (PnP) algorithms \cite{peng2019pvnet,pavlakos20176,rad2017bb8,tekin2018real}. Although these two-stage approaches performed more stable,  most of them were built on top of the 2D projection. Errors that are small in projection can be large in real 3D space. Also, different keypoints in 3D space may be overlapped after 2D projection, making them hard to be distinguished. Moreover, geometric constraint information of rigid objects would be partially lost due to projection.

On the other hand,  with the development of inexpensive RGBD sensors, more and more RGBD datasets are available. The extra depth information allows 2D algorithms to be extend into 3D space with better performance, like PointFusion \cite{xu2018pointfusion}, Frustum pointnets\cite{qi2018frustum} and VoteNet\cite{qi2019deep}. Towards this end, we extend 2D-keypoint-based approaches to 3D keypoint to fully utilize geometric constraint information of rigid objects and significantly improved the accuracy of 6DoF estimation.  More specifically, we develop a deep 3D keypoints Hough voting neural network to learn the point-wise 3D offset and vote for 3D keypoints, as shown in Figure \ref{fig:our_approach}.  Our key observation is a simple geometric property that positional relationship between two points of a rigid object in 3D space is fixed. Hence, given a visible point on the object surface, its coordinate and orientation can be obtained from depth images and its translation offset to selected keypoint is also fixed and learnable. Meanwhile, learning point-wise Euclidean offset is straightforward for network and easier to optimize. 

To handle scenes with multiple objects, we also introduce an instance semantic segmentation module into the network and jointly optimized with keypoint voting. We find that jointly training these tasks boosts the performance of each other. Specifically, semantic information improves translation offset learning by identifying which part a point belongs to  and the size information contained in translation offsets helps the model to distinguish objects with similar appearance but different size.


We further conduct experiments on YCB-Video and LineMOD datasets to evaluate our method. Experimental results show that our approach outperforms current state-of-the-art methods by a significant margin.

To summarize, the main contributions of this work are as follows:
\begin{itemize}
\item A novel deep 3D keypoints Hough voting network with instance semantic segmentation for 6DoF Pose Estimation of single RGBD image.
\item State-of-the-art 6DoF pose estimation performance on YCB and LineMOD datasets.
\item An in-depth analysis of our 3D-keypoint-based method and comparison with previous approaches, demonstrating that 3D-keypoint is a key factor to boost performance for 6DoF pose estimation. We also show that jointly training 3D-keypoint and semantic segmentation can further improve the performance.
\end{itemize}

%% file: related_work.tex
\section{Related Work}
\subsection{Holistic Methods}
Holistic methods directly estimate the 3D position and orientation of objects in a given image. Classical template-based methods construct rigid templates and scan through the image to compute the best matched pose \cite{huttenlocher1993comparing,gu2010discriminative,hinterstoisser2011gradient}. Such templates are not robust to clustered scenes. Recently, some Deep Neural Network (DNN) based methods are proposed to directly regress the 6D pose of cameras or objects \cite{xiang2017posecnn,wang2019densefusion,gupta2019cullnet}. However, non-linearity of the rotation space making the data-driven DNN hard to learn and generalize. To address this problem, some approaches use post-refinement procedure \cite{li2018deepim,wang2019densefusion} to refine the pose iteratively, others discrete the rotation space and simplify it to be a classification problem \cite{tulsiani2015viewpoints,su2015render,sundermeyer2018implicit}. For the latter approach, post-refinement processes are still required to compensate for the accuracy sacrificed by the discretization. 

\subsection{Keypoint-based Methods}
Current keypoint-based methods first detect the 2D keypoints of an object in the images, then utilize a PnP algorithm to estimate the 6D pose. Classical methods \cite{lowe1999object,rothganger20063d,bay2006surf} are able to detect 2D keypoint of objects with rich texture efficiently. However, they can not handle texture-less objects. With the development of deep learning techniques, some neural-network-based 2D keypoints detection methods are proposed. \cite{rad2017bb8,tekin2018real,hu2019segmentation} directly regress the 2D coordinate of the keypoints, while \cite{newell2016stacked,kendall2015posenet,oberweger2018making} use heatmaps to locate the 2D keypoints. To better deal with truncated and occluded scenes, \cite{peng2019pvnet} proposes a pixel-wise voting network to vote for the 2D keypoints location. These 2D keypoint based methods aim to minimize the 2D projection errors of objects. However, errors that are small in projection may be large in the real 3D world. \cite{keypointnet} extracts 3D keypoints from two views of synthetic RGB images to recover 3D poses. Nevertheless, they only utilize the RGB images, on which geometric constraint information of rigid objects partly lost due to projection, and different keypoints in 3D space may be overlapped and hard to be distinguished after projected to 2D. The advent of cheap RGBD sensors enables us to do everything in 3D with captured depth images. 

\subsection{Dense Correspondence Methods}
These approach utilize Hough voting scheme \cite{liebelt2008independent,sun2010depth,glasner2011aware} to vote for final results with per-pixel prediction. They either use random forest \cite{brachmann2014learning,michel2017global} or CNNs \cite{kehl2016deep,doumanoglou2016recovering,Li_2019_ICCV,park2019pix2pose,wang2019normalized} to extract feature and predict the corresponding 3D object coordinates for each pixel and then vote for the final pose results. Such dense 2D-3D correspondence making these methods robust to occluded scenes, while the output space is quite large. PVNet \cite{peng2019pvnet} uses per-pixel voting for 2D Keypoints to combine the advantages of Dense methods and keypoint-based methods. We further extend this method to 3D keypoints with extra depth information and fully utilize geometric constraints of rigid objects.


%% file: method.tex
\begin{figure*}
    \centering
    \includegraphics[scale=0.145]{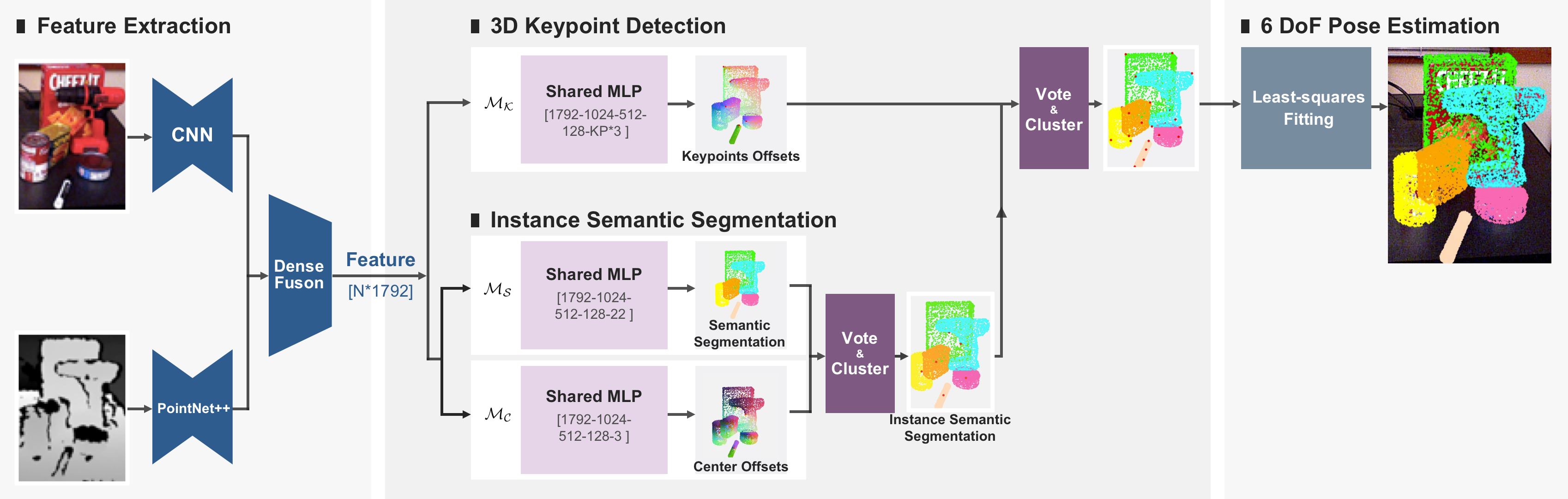}
    \caption{
        \textbf{Overview of PVN3D.} The Feature Extraction module extracts the per-point feature from an RGBD image. They are fed into module $\mathcal{M_K}$, $\mathcal{M_C}$ and $\mathcal{M_S}$ to predict the translation offsets to keypoints, center point and semantic labels of each point respectively. A clustering algorithm is then applied to distinguish different instances with the same semantic label and points on the same instance vote for their target keypoints. Finally, a least-square fitting algorithm is applied to the predicted keypoints to estimate 6DoF pose parameters. 
    }
    \label{fig:network}
\end{figure*}

\section{Proposed Method}

Given an RGBD image, the task of 6DoF pose estimation is to estimate the rigid transformation that transforms an object from its object world coordinate system to the camera world coordinate system. Such transformation consists of a 3D rotation $\mathbf{R} \in SO(3)$ and a translation $\mathbf{t} \in \mathbb{R}^3$.

\subsection{Overview}


To tackle this task, we develop a novel approach based on a deep 3D Hough voting network, as shown in Figure \ref{fig:network}.  The proposed method is a two-stage pipeline with 3D keypoint detection followed by a pose parameters fitting module. More specifically, taking an RGBD image as input, a feature extraction module would be used to fuse the appearance feature and geometry information. The learned feature would be fed into a 3D keypoint detection module $\mathbfcal{M_K}$ which was trained to predict the per-point offsets w.r.t keypoints. Additionally, we include an instance segmentation module for multiple objects handling where a semantic segmentation module $\mathbfcal{M_S}$ predicts the per-point semantic label, and a center voting module $\mathbfcal{M_C}$ predicts the per-point offsets to object center.  With the learned per-point offset, the clustering algorithm \cite{comaniciu2002mean} is applied to distinguish different instances with the same semantic label and points on the same instance vote for their target keypoints. Finally, a least-square fitting algorithm is applied to the predicted keypoints to estimate 6DoF pose parameters.

\subsection{Learning Algorithm}


The goal of our learning algorithm is to train a 3D keypoint detection module $\mathbfcal{M_K}$ for offset prediction as well as a semantic segmentation module $\mathbfcal{M_S}$ and center voting module $\mathbfcal{M_C}$ for instance-level segmentation.  This naturally makes training our network multi-task learning, which is achieved by a supervised loss we designed and several training details we adopt.  

{\textbf{3D Keypoints Detection Module.}} As shown in Figure \ref{fig:network}, with the per-point feature extracted by the feature extraction module, a 3D keypoint detection module $\mathcal{M_K}$ is used to detect the 3D keypoints of each object. To be specific, $\mathbfcal{M_K}$ predicts the per-point Euclidean translation offset from visible points to target keypoints. These visible points, together with the predicted offsets then vote for the target keypoints. The voted points are then gathered by clustering algorithms and centers of clusters are selected as the voted keypoints.

We give a deeper view of $\mathbfcal{M_K}$ as follows. Given a set of visible seed points $\{p_i\}^N_{i=1}$ and a set of selected keypoints $\{kp_j\}^M_{j=1}$ belonging to the same object instance $I$, we denote $p_i = [x_i; f_i]$ with $x_i$ the 3D coordinate and $f_i$ the extracted feature. We denote $kp_j = [y_j]$ with $y_j$ the 3D coordinate of the keypoint. $\mathbfcal{M_K}$ absorbs feature $f_i$ of each seed point and generates translation offset $\{of_i^j\}_{j=1}^M$ for them, where $of_i^j$ denotes the translation offset from the $i_{th}$ seed point to the $j_{th}$ keypoint. Then the voted keypoint can be denoted as $vkp_i^{j} = x_i + of_i^{j}$. To supervise the learning of $of_i^j$, we apply an L1 loss:
\begin{equation}\label{eqn:Lofs}
    L_{\textrm{keypoints}} = \frac{1}{N} \sum_{i=1}^{N} \sum_{j=1}^M ||of_i^j - of_i^{j*}|| \mathbb{I}(p_i \in I)
\end{equation}
where $of_i^{j*}$ is the ground truth translation offset; $M$ is the total number of selected target keypoints; $N$ is the total number of seeds and $\mathbb{I}$ is an indicating function equates to $1$ only when point $p_i$ belongs to instance $I$, and $0$ otherwise.

{\textbf{Instance Semantic Segmentation Module.}}
 To handle scenes with multi objects, previous methods \cite{wang2019densefusion, xu2018pointfusion, qi2018frustum} utilize existing detection or semantic segmentation architecture to pre-process the image and obtain RoIs (regions of interest) containing only single objects. Then build the pose estimation models with the extracted ROIs as input to simplify the problem. However, as we have formulated the pose estimation problem to first detect keypoints of objects with a translation offsets to keypoints learning module, we believe that the two tasks can enhance the performance of each other. On the one hand, the semantic segmentation module forces the model to extract global and local features on instance to distinguish different objects, which helps to locate a point on the object and does good to the keypoint offset reasoning procedure. On the other hand, size information learned for the prediction of offsets to the keypoints helps distinguish objects with similar appearance but different in size. Under such observation, we introduce a point-wise instance semantic segmentation module $\mathcal{M_S}$ into the network and jointly optimized it with module $\mathcal{M_K}$. 
 
 To be specific, given the per-point extracted feature, the semantic segmentation module $\mathbfcal{M_S}$ predicts the per-point semantic labels. We supervise this module with Focal Loss \cite{lin2017focal}:
 \begin{equation}
     \label{eqn:L_sem}
     \begin{split}
     L_{semantic} = & -\alpha(1-q_i)^\gamma log(q_i) \\
                    & where \quad q_i = c_i \cdot l_i
     \end{split}
 \end{equation}
with $\alpha$ the $\alpha$-balance parameter, $\gamma$ the focusing parameter, $c_i$ the predicted confidence for the $i_{th}$ point belongs to each class and $l_i$ the one-hot representation of ground true class label.
  
 Meanwhile, the center voting module $\mathbfcal{M_C}$ is applied to vote for centers of different object so as to distinguish different instance. We propose such module under the inspiration of CenterNet \cite{duan2019centernet} but further extend the 2D center point to 3D. Compared to 2D center points, different center points in 3D won't suffer from occlusion due to camera projection in some viewpoints.  Since we can regard the center point as a special keypoint of an object, module $\mathbfcal{M_C}$ is similar to the 3D keypoint detection module $\mathbfcal{M_K}$. It takes in the per-point feature but predicts the Euclidean translation offset $\Delta x_i$ to the center of objects it belongs to. The learning of $\Delta x_i$ is also supervised by an L1 loss:
\begin{equation}
\label{eqn:Lctr}
    L_{\textrm{center}} = \frac{1}{N} \sum_{i=1}^{N}{ ||\Delta x_i - \Delta x_i^* || \mathbb{I}(p_i \in I)
    }
\end{equation}
where $N$ denotes the total number of seed points on the object surface and $\Delta x_i^*$ is the ground truth translation offset from seed $p_i$ to the instance center. $\mathbb{I}$ is an indication function indicating whether point $p_i$ belongs to that instance.


{\textbf{Multi-task loss.}} We supervise the learning of module $\mathbfcal{M_K}$, $\mathbfcal{M_S}$ and $\mathbfcal{M_C}$ jointly with a multi-tasks loss:
\begin{equation}\label{eqn:L}
    \begin{split}
    L_{\textrm{multi-task}} = & \lambda_1 L_{\textrm{keypoints}} + \lambda_2 L_{\textrm{semantic}} + \lambda_3 L_{\textrm{center}}
    \end{split}
\end{equation}
where $\lambda_1$, $\lambda_2$ and $\lambda_3$ are the weights for each task. Experimental results shows that jointly training these tasks boosts the performance of each other.

\subsection{Training and Implementation}

\renewcommand{\arraystretch}{1.5}
\newcommand{\ycbC}{0.7}
\begin{table*}[tp]
    \centering
    \fontsize{7.2}{7.5}\selectfont
    \begin{tabular}{l|C{\ycbC cm}|C{\ycbC cm}|C{\ycbC cm}|C{\ycbC cm}|C{\ycbC cm}|C{\ycbC cm}|C{\ycbC cm}|C{\ycbC cm}|C{\ycbC cm}|C{\ycbC cm}|C{\ycbC cm}|C{\ycbC cm}}
    \hline
                                  & \multicolumn{6}{c|}{\begin{tabular}[c]{@{}c@{}}Without Iterative Refinement\end{tabular}}       & \multicolumn{6}{c}{With Iterative Refinement}                                                           \cr\hline
                                  & \multicolumn{2}{c|}{PoseCNN\cite{xiang2017posecnn}}  & \multicolumn{2}{c|}{DF(per-pixel)\cite{wang2019densefusion}} & \multicolumn{2}{c|}{PVN3D}   & \multicolumn{2}{c|}{PoseCNN+ICP\cite{xiang2017posecnn}} & \multicolumn{2}{c|}{DF(iterative)\cite{wang2019densefusion}} & \multicolumn{2}{c}{PVN3D+ICP} \cr\hline
                                  & ADDS         & ADD(S)        & ADDS           & ADD(S)           & ADDS        & ADD(S)        & ADDS          & ADD(S)          & ADDS           & ADD(S)           & ADDS         & ADD(S)         \cr\hline
        002\_master\_chef\_can            & 83.9        & 50.2          & 95.3            & 70.7            & 96.0              & \textbf{80.5}     & 95.8           & 68.1           & \textbf{96.4}   & 73.2            & 95.2          & 79.3          \\
        003\_cracker\_box                 & 76.9        & 53.1          & 92.5            & 86.9            & \textbf{96.1}     & \textbf{94.8}     & 92.7           & 83.4           & 95.8            & 94.1            & 94.4          & 91.5          \\
        004\_sugar\_box                   & 84.2        & 68.4          & 95.1            & 90.8            & 97.4              & 96.3              & \textbf{98.2}  & \textbf{97.1}  & 97.6            & 96.5            & 97.9          & 96.9          \\
        005\_tomato\_soup\_can            & 81.0        & 66.2          & 93.8            & 84.7            & \textbf{96.2}     & 88.5              & 94.5           & 81.8           & 94.5            & 85.5            & 95.9          & \textbf{89.0} \\
        006\_mustard\_bottle              & 90.4        & 81.0          & 95.8            & 90.9            & 97.5              & 96.2              & \textbf{98.6}  & \textbf{98.0}  & 97.3            & 94.7            & 98.3          & 97.9          \\
        007\_tuna\_fish\_can              & 88.0        & 70.7          & 95.7            & 79.6            & 96.0              & 89.3              & \textbf{97.1}  & 83.9           & \textbf{97.1}   & 81.9            & 96.7          & \textbf{90.7} \\
        008\_pudding\_box                 & 79.1        & 62.7          & 94.3            & 89.3            & 97.1              & 95.7              & 97.9           & 96.6           & 96.0            & 93.3            & \textbf{98.2} & \textbf{97.1} \\
        009\_gelatin\_box                 & 87.2        & 75.2          & 97.2            & 95.8            & 97.7              & 96.1              & 98.8           & 98.1           & 98.0            & 96.7            & \textbf{98.8} & \textbf{98.3} \\
        010\_potted\_meat\_can            & 78.5        & 59.5          & 89.3            & 79.6            & 93.3              & \textbf{88.6}     & 92.7           & 83.5           & 90.7            & 83.6            & \textbf{93.8} & 87.9          \\
        011\_banana                       & 86.0        & 72.3          & 90.0            & 76.7            & 96.6              & 93.7              & 97.1           & 91.9           & 96.2            & 83.3            & \textbf{98.2} & \textbf{96.0} \\
        019\_pitcher\_base                & 77.0        & 53.3          & 93.6            & 87.1            & 97.4              & 96.5              & \textbf{97.8}  & 96.9           & 97.5            & 96.9            & 97.6          & \textbf{96.9} \\
        021\_bleach\_cleanser             & 71.6        & 50.3          & 94.4            & 87.5            & 96.0              & 93.2              & 96.9           & 92.5           & 95.9            & 89.9            & \textbf{97.2} & \textbf{95.9} \\
        \textbf{024\_bowl}                & 69.6        & 69.6          & 86.0            & 86.0            & 90.2              & 90.2              & 81.0           & 81.0           & 89.5            & 89.5            & \textbf{92.8} & \textbf{92.8} \\
        025\_mug                          & 78.2        & 58.5          & 95.3            & 83.8            & 97.6              & 95.4              & 94.9           & 81.1           & 96.7            & 88.9            & \textbf{97.7} & \textbf{96.0} \\
        035\_power\_drill                 & 72.7        & 55.3          & 92.1            & 83.7            & 96.7              & 95.1              & \textbf{98.2}  & \textbf{97.7}  & 96.0            & 92.7            & 97.1          & 95.7          \\
        \textbf{036\_wood\_block}         & 64.3        & 64.3          & 89.5            & 89.5            & 90.4              & 90.4              & 87.6           & 87.6           & \textbf{92.8}   & \textbf{92.8}   & 91.1          & 91.1          \\
        037\_scissors                     & 56.9        & 35.8          & 90.1            & 77.4            & \textbf{96.7}     & \textbf{92.7}     & 91.7           & 78.4           & 92.0            & 77.9            & 95.0          & 87.2          \\
        040\_large\_marker                & 71.7        & 58.3          & 95.1            & 89.1            & 96.7              & 91.8              & 97.2           & 85.3           & 97.6            & \textbf{93.0}   & \textbf{98.1} & 91.6          \\
        \textbf{051\_large\_clamp}        & 50.2        & 50.2          & 71.5            & 71.5            & 93.6              & 93.6              & 75.2           & 75.2           & 72.5            & 72.5            & \textbf{95.6} & \textbf{95.6} \\
        \textbf{052\_extra\_large\_clamp} & 44.1        & 44.1          & 70.2            & 70.2            & 88.4              & 88.4              & 64.4           & 64.4           & 69.9            & 69.9            & \textbf{90.5} & \textbf{90.5} \\
        \textbf{061\_foam\_brick}         & 88.0        & 88.0          & 92.2            & 92.2            & 96.8              & 96.8              & 97.2           & 97.2           & 92.0            & 92.0            & \textbf{98.2} & \textbf{98.2} \cr\hline
        ALL                               & 75.8        & 59.9          & 91.2            & 82.9            & 95.5              & 91.8              & 93.0           & 85.4           & 93.2            & 86.1            & \textbf{96.1} & \textbf{92.3} \cr\hline
    \end{tabular}
    \caption{Quantitative evaluation of 6D Pose (ADD-S AUC \cite{xiang2017posecnn}, ADD(S) AUC \cite{hinterstoisser2012model}) on the YCB-Video Dataset. Symmetric objects' names are in bold.
    }
    \label{tab:YCB_PFM}
\end{table*}

\renewcommand{\arraystretch}{1.5}
\begin{table}[tp]
    \centering
    \fontsize{6.4}{6.4}\selectfont
    
    \begin{tabular}{l|L{0.75cm}|C{0.75cm}|C{0.75cm}|C{0.75cm}|C{1.02cm}}
    \hline
                                          &        & \multicolumn{2}{c|}{\begin{tabular}[c]{@{}l@{}}w/o iter. ref. \end{tabular}} & \multicolumn{2}{c}{w/ iter. ref.} \cr\hline
                                          &        &DF(p.p.)                                   &PVN3D                                  &DF(iter.)       &PVN3D+ICP           \cr\hline
        \textbf{large \_clamp}        & ADD-S  & 87.7                                          & 93.9                                   & 90.3                & \textbf{96.2}       \\
        \textbf{extra \_large\_clamp} & ADD-S  & 75.0                                          & 90.1                                   & 74.9                & \textbf{93.6}       \cr\hline
        \multirow{2}{*}{ALL}          & ADD-S  & 93.3                                          & 95.7                                   & 94.8                & \textbf{96.4}       \\
                                      & ADD(S) & 84.9                                          & 91.9                                   & 89.4                & \textbf{92.7}      \cr\hline
    \end{tabular}

    \caption{Quantitative evaluation results on the YCB-Video dataset with ground truth instance semantic segmentation result.}
    \label{tab:YCB_PF_GT}
\end{table}

{\textbf{Network Architecture.}}
 The first part in Figure \ref{fig:network} is a feature extraction module. In this module, a PSPNet \cite{zhao2017pyramid} with an ImageNet \cite{deng2009imagenet} pretrained ResNet34 \cite{resnet} is applied to extract the appearance information in RGB images. A PointNet++ \cite{qi2017pointnet++} extracts the geometry information in point clouds and their normal maps. They are further fused by a DenseFusion block \cite{wang2019densefusion} to obtain the combined feature for each point. After the process of this module, each point $p_i$ has a feature $f_i \in \mathbb{R}^C$ of $C$ dimension. The following module $\mathbfcal{M_K}$, $\mathbfcal{M_S}$ and $\mathbfcal{M_C}$ are composed of shared Multi-Layer Perceptrons (MLPs) shown in Figure \ref{fig:network}. We sample $N=12288$ points (pixels) for each frame of RGBD image and set $\lambda_1 = \lambda_2 = \lambda_3 = 1.0$ in Formula \ref{eqn:L}. 

{\textbf{Keypoint Selection.}} The 3D keypoints are selected from 3D object models. In 3D object detection algorithms \cite{qi2018frustum,xu2018pointfusion,qi2019deep}, eight corners of the 3D bounding box are selected. However, These bounding box corners are virtual points that are far away from points on the object, making point-based networks difficult to aggregate scene context in the vicinity of them. The longer distance to the object points results in larger localization errors, which may do harm to the compute of 6D pose parameters. Instead, points selected from the object surface will be quite better. Therefore, we follow \cite{peng2019pvnet} and use the farthest point sampling (FPS) algorithm to select keypoints on the mesh. Specifically, we initial the selection procedure by adding the center point of the object model in an empty keypoint set. Then update it by adding a new point on the mesh that is farthest to all the selected keypoints repeatedly, until $M$ keypoints are obtained.

{\textbf{Least-Squares Fitting.}}
Given two point sets of an object, one from the $M$ detected keypoints $\{kp_j\}_{j=1}^{M}$ in the camera coordinate system, and another from their corresponding points $\{kp_j^{'}\}_{j=1}^{M}$ in the object coordinate system, the 6D pose estimation module computes the pose parameters ($R$, $t$) with a least-squares fitting algorithm \cite{arun1987least}, which finds $R$ and $t$ by minimizing the following square loss:
\begin{equation} \label{eqn:lsf}
    L_{\textrm{least-squares}} = \sum_{j=1}^{M} || kp_j - (R \cdot kp^{'}_j + t) ||^2
\end{equation}
where $M$ is the number of selected keypoints of a object.

%% file: results.tex
\section{Experiments}
\subsection{Datasets}
We evaluate our method on two benchmark datasets. 

\textbf{YCB-Video Dataset} contains 21 YCB \cite{calli2015ycb} objects of varying shape and texture. 92 RGBD videos of the subset of objects were captured and annotated with 6D pose and instance semantic mask. The varying lighting conditions, significant image noise, and occlusions make this dataset challenging. We follow \cite{xiang2017posecnn} and split the dataset into 80 videos for training and another 2,949 keyframes chosen from the rest 12 videos for testing. Following \cite{xiang2017posecnn}, we add the synthetic images into our training set. A hole completion algorithm \cite{ku2018defense} is also applied to improve the quality of depth images. 

\textbf{LineMOD Dataset} \cite{hinterstoisser2011multimodal} consists of 13 low-textured objects in 13 videos, annotated 6D pose and instance mask. The main challenge of this dataset is the cluttered scenes, texture-less objects, and lighting variations. We follow prior works \cite{xiang2017posecnn} to split the training and testing set. Also, we follow \cite{peng2019pvnet} and add synthesis images into our training set.

\renewcommand{\arraystretch}{1.3}
\begin{table*}[tp]
    \centering
    \fontsize{7.0}{6.8}\selectfont
    \begin{tabular}{l|C{1.1cm}|C{1.1cm}|C{1.1cm}|C{1.1cm}|C{1.1cm}|C{1.1cm}|C{1.1cm}|C{1.1cm}|C{1.1cm} }
        \hline
         \  & \multicolumn{3}{c|}{RGB}&\multicolumn{6}{c}{RGBD}\cr\cline{1-10}
         \  &PoseCNN DeepIM \cite{li2018deepim,xiang2017posecnn}	&PVNet \cite{peng2019pvnet} &CDPN \cite{Li_2019_ICCV} &Implicit ICP\cite{sundermeyer2018implicit} &SSD-6D ICP\cite{kehl2017ssd} &Point-Fusion\cite{wang2019densefusion}	&DF(per-pixel)\cite{wang2019densefusion}	&DF(ite-rative)\cite{wang2019densefusion}	&PVN3D \cr
        \hline
        ape         & 77.0          & 43.6          & 64.4 & 20.6 & 65.0           & 70.4 & 79.5 & 92.3           & \textbf{97.3}  \\
        benchvise   & 97.5          & \textbf{99.9} & 97.8 & 64.3 & 80.0           & 80.7 & 84.2 & 93.2           & 99.7           \\
        camera      & 93.5          & 86.9          & 91.7 & 63.2 & 78.0           & 60.8 & 76.5 & 94.4  & \textbf{99.6}           \\
        can         & 96.5 & 95.5          & 95.9 & 76.1 & 86.0           & 61.1 & 86.6 & 93.1           & \textbf{99.5}           \\
        cat         & 82.1          & 79.3          & 83.8 & 72.0 & 70.0           & 79.1 & 88.8 & 96.5  & \textbf{99.8}           \\
        driller     & 95.0          & 96.4 & 96.2 & 41.6 & 73.0           & 47.3 & 77.7 & 87.0           & \textbf{99.3}           \\
        duck        & 77.7          & 52.6          & 66.8 & 32.4 & 66.0           & 63.0 & 76.3 & 92.3           & \textbf{98.2}  \\
        \textbf{eggbox}      & 97.1          & 99.2          & 99.7 & 98.6 & \textbf{100.0} & 99.9 & 99.9 & 99.8           & 99.8 \\
        \textbf{glue}        & 99.4          & 95.7          & 99.6 & 96.4 & \textbf{100.0} & 99.3 & 99.4 & \textbf{100.0} & \textbf{100.0} \\
        holepuncher & 52.8          & 82.0          & 85.8 & 49.9 & 49.0           & 71.8 & 79.0 & 92.1           & \textbf{99.9}  \\
        iron        & 98.3          & 98.9 & 97.9 & 63.1 & 78.0           & 83.2 & 92.1 & 97.0           & \textbf{99.7}           \\
        lamp        & 97.5          & 99.3 & 97.9 & 91.7 & 73.0           & 62.3 & 92.3 & 95.3           & \textbf{99.8}           \\
        phone       & 87.7          & 92.4          & 90.8 & 71.0 & 79.0           & 78.8 & 88.0 & 92.8           & \textbf{99.5}  \cr\hline
        ALL         & 88.6          & 86.3          & 89.9 & 64.7 & 79.0           & 73.7 & 86.2 & 94.3           & \textbf{99.4} \cr
        \hline  
    \end{tabular}
    \caption{Quantitative evaluation of 6D Pose on ADD(S) \cite{hinterstoisser2012model} metric on the LineMOD dataset. Objects with bold name are symmetric.}
    \label{tab:LM_PFM}
\end{table*}

\newcommand{\as}{1.57}
\begin{table*}[tp]
    \centering
    \fontsize{7.0}{6.8}\selectfont
    \begin{tabular}{l|C{\as cm}|C{\as cm}|C{\as cm}|C{\as cm}|C{\as cm}|C{\as cm}|C{\as cm}|C{\as cm}}
        \hline
         \      & DF(RT)\cite{wang2019densefusion} & DF(3D KP)\cite{wang2019densefusion} &Ours(RT) &Ours(2D KPC) &Ours(2D KP) &PVNet\cite{peng2019pvnet}  &Ours(Corr) &Ours(3D KP)\cr\cline{1-9}
        ADD-S	&92.2	 &93.1	     &92.8	   &78.2           &81.8	    &-      &92.8	    &{\bf95.5}   \cr
        ADD(S)	&86.9	 &87.9	     &87.3	   &73.8           &77.2        &73.4	&88.1	    &{\bf91.8}   \cr
        \hline
    \end{tabular}
    \caption{Quantitative evaluation of 6D Poses on the YCB-Video dataset with different formulations. All with our predicted segmentation.}
    \label{tab:YCB_FM}
\end{table*}

\newcommand{\kpC}{0.95}
\begin{table}[tp]
    \centering
    \fontsize{6.9}{6.8}\selectfont
    \begin{tabular}{l|C{1.05 cm}|C{\kpC cm}|C{\kpC cm}|C{\kpC cm}|C{\kpC cm} }
        \hline
        \       &VoteNet\cite{qi2019deep} &BBox 8 &FPS 4  &FPS 8      &FPS 12 \cr\hline
        ADD-S   &89.9    &94.0   &94.3	 &{\bf95.5}  &94.5   \cr
        ADD(S)  &85.1    &90.2   &90.5	 &{\bf91.8}  &90.7   \cr
        \hline  
    \end{tabular}
    \caption{Effect of different keypoint selection methods of PVN3D. Results of VoteNet\cite{qi2019deep}, another 3D bounding box detection approach are added as a simple baseline to compare with our BBox8.}
    \label{tab:KP_SL}
\end{table}

\subsection{Evaluation Metrics}
We follow \cite{xiang2017posecnn} and evaluate our method with the average distance ADD and ADD-S metric \cite{xiang2017posecnn}. The average distance ADD metric \cite{hinterstoisser2012model} evaluates the mean pair-wise distance between object vertexes transformed by the predicted 6D pose [$R$, $t$] and the ground true pose [$R^*$, $t^*$]:
\begin{equation}
    \label{eqn:ADD}
    \textrm{ADD} = \frac{1}{m} \sum_{x \in \mathcal{O}} || (Rx+t) - (R^*x + t^*) ||
\end{equation}
where $x$ is a vertex of totally $m$ vertexes on the object mesh $\mathcal{O}$. The ADD-S metric is designed for symmetric objects and the mean distance is computed based on the closest point distance:
\begin{equation}
    \label{eqn:ADDS}
    \textrm{ADD{-}S} = \frac{1}{m} \sum_{x_1 \in \mathcal{O}} \min_{x_2 \in \mathcal{O}}{|| (Rx_1+t) - (R^*x_2 + t^*) ||}
\end{equation}
For evaluation, we follow \cite{xiang2017posecnn,wang2019densefusion} and compute the ADD-S AUC, the area under the accuracy-threshold curve, which is obtained by varying the distance threshold in evaluation. The ADD(S)\cite{hinterstoisser2012model} AUC is computed in a similar way but calculate ADD distance for non-symmetric objects and ADD-S distance for symmetric objects.

\newcommand{\gwCCMP}{15.4}
\begin{figure*}[htbp]
\centering
\includegraphics[width=\gwCCMP cm]{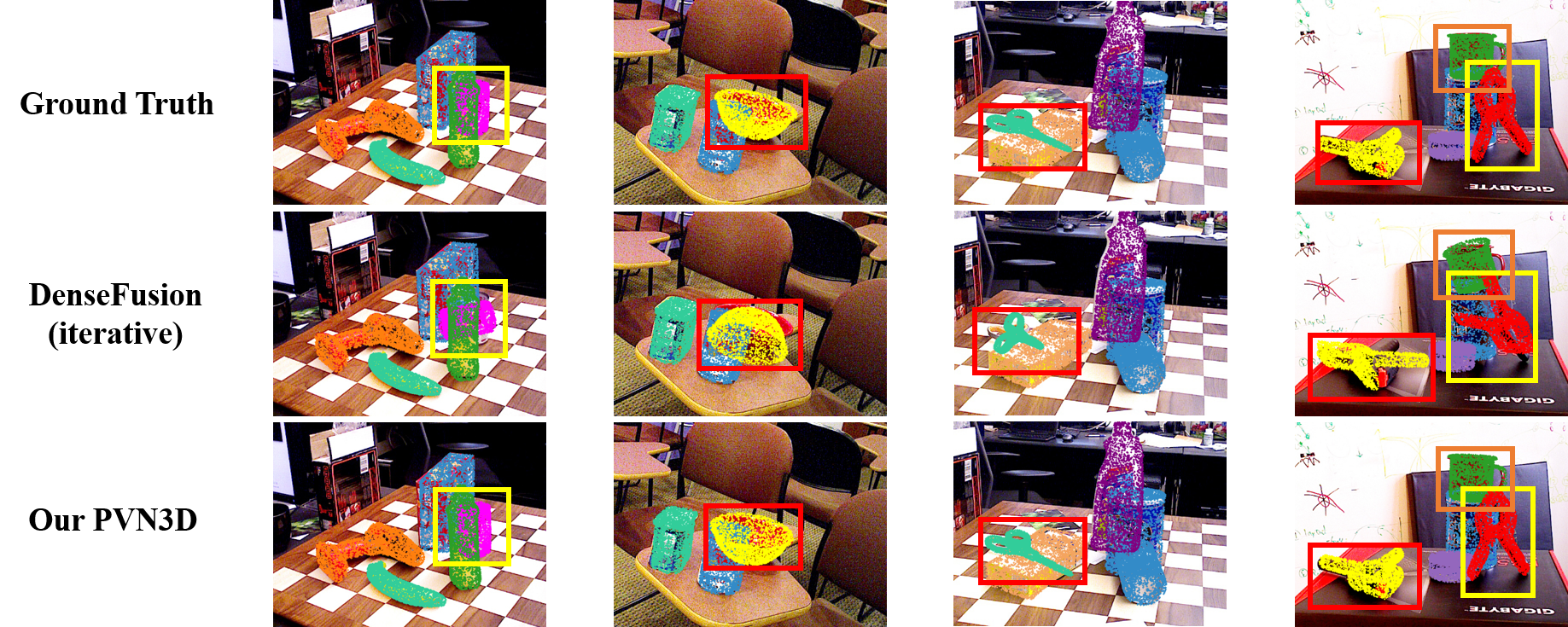}
\centering
\caption{
    \textbf{Qualitative results on the YCB-Video dataset.} Points on different meshes in the same scene are in different colors. They are projected back to the image after being transformed by the predicted pose. We compare our PVN3D \textbf{without any iterative refinement procedure} to DenseFusion with iterative refinement (2 iterations). Our model distinguishes the challenging large clamp and extra-large clamp and estimates their poses well. Our model is also robust in heavily occluded scenes. 
}
\vspace{-0.2cm}
\label{fig:compare_fig}
\end{figure*}

\begin{figure}
    \centering
    \includegraphics[scale=0.50]{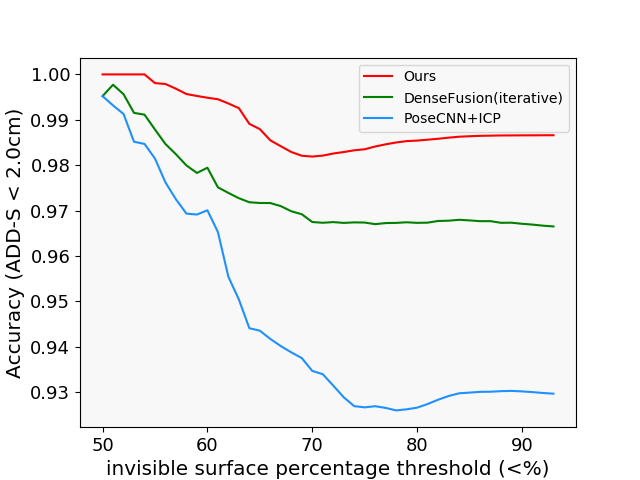}
    \caption{
        Performance of different approaches under increasing levels of occlusion on the YCB-Video dataset.
    }
    \label{fig:occlussion_auc}
\end{figure}

\subsection{Evaluation on YCB-Video \& LineMOD Dataset}
Table \ref{tab:YCB_PFM} shows the evaluation results for all the 21 objects in the YCB-Video dataset. We compare our model with other single view methods. As shown in the Table, our model without any iterative refinement procedure (PVN3D) surpasses all other approaches by a large margin, even when they are iterative refined. On the ADD(S) metric, our model outperforms PoseCNN+ICP \cite{xiang2017posecnn} by 6.4\% and exceeds DF(iterative) \cite{wang2019densefusion} by 5.7\%. With iterative refinement, our model (PVN3D+ICP) achieves even better performance. Note that one challenge of this dataset is to distinguish the large clamp and extra-large clamp, on which previous methods \cite{wang2019densefusion, xiang2017posecnn} suffer from poor detection results. We also report evaluation results with ground truth segmentation in Table \ref{tab:YCB_PF_GT}, which shows that our PVN3D still achieves the best performance. Some Qualitative results are shown in Figure \ref{fig:compare_fig}. Table \ref{tab:LM_PFM} demonstrates the evaluation results on LineMOD dataset. Our model also achieves the best performance.

{\textbf{Robust to Occlusion Scenes.}} One of the biggest advantages of our 3D-keypoint-based method is that it's robust to occlusion naturally. To explored how different methods are influenced by different degrees of occlusion, we follow \cite{wang2019densefusion} and calculate the percentage of invisible points on the object surface. Accuracy of ADD-S $<$ 2cm under different invisible surface percentage is shown in Figure \ref{fig:occlussion_auc}. The performance of different approaches is very close when 50\% of points are invisible. However, with the percentage of invisible part increase, DenseFusion and PoseCNN+ICP fall faster comparing with ours. Figure \ref{fig:compare_fig} shows that our model performs well even when objects are heavily occluded.

\subsection{Ablation Study}
In this part, we explore the influence of different formulation for 6DoF pose estimation and the effect of keypoint selection methods. We also probe the effect of multi-task learning.

{\textbf{Comparisons to Directly Regressing Pose.}} To compare our 3D keypoint based formulation with formulations that directly regressing the 6D pose parameters [$R$, $t$] of an object, we simply modify our 3D keypoint voting module $\mathbfcal{M_K}$ to directly regress the quaternion rotation $R$ and the translation parameters $t$ for each point. We also add a confidence header following DenseFusion \cite{wang2019densefusion} and select the pose with the highest confidence as the final proposed pose. We supervise the training process using ShapeMatch-Loss \cite{xiang2017posecnn} with confidence regularization term \cite{wang2019densefusion} following DenseFusion. Experimental results in Table \ref{tab:YCB_FM} shows that our 3D keypoint formulation performs quite better. 

To eliminate the influence of different network architecture, we also modify the header of DenseFusion(per-pixel) to predict the per-point translation offset and compute the 6D pose following our keypoint voting and least-squares fitting procedure. Table \ref{tab:YCB_FM} reveals that the 3D keypoint formulation, DF(3D KP) in the Table, performs better than the RT regression formulation, DF(RT). That's because the 3D keypoint offset search space is smaller than the non-linearity of rotation space, which is easier for neural networks to learn, enabling them to be more generalizable. 

{\textbf{Comparisons to 2D Keypoints.}} In order to contrast the influence of 2D and 3D keypoints, we project the voted 3D keypoints back to 2D with the camera intrinsic parameters. A PnP algorithm with Random Sample Consensus (RANSAC) is then applied to compute the 6D pose parameters. Table \ref{tab:YCB_FM} shows that algorithms with 3D keypoint formulation, denoted as Ours(3D KP) in the table, outperforms 2D keypoint, denoted Ours(2D KP) in the table,  by 13.7\% under ADD-S metric. 
That's because PnP algorithms aim to minimize the projection error. However, pose estimation errors that are small in projection may be quite large in the 3D real world.

To compare the influence between 2D and 3D center point in our instance semantic segmentation module, we also project our voted 3D center point to 2D in the Instance Semantic Segmentation module(Ours(2D KPC)). We apply a similar Mean Shift algorithm to cluster the voted 2D center points to distinguish different instance, finding that in occlusion scenes, different instances are hard to be differentiated when their centers are close to each other after projected on 2D, while they are far away from each other and can be easily differentiated in 3D real world. Note that other existing 2D keypoints detection approaches, such as heat-map \cite{newell2016stacked,kendall2015posenet,oberweger2018making} and vector voting \cite{peng2019pvnet} models may also suffer from overlapped keypoints. By definition, centers of most objects in our daily life won't be overlapped as they usually lie within the object while they may be overlapped after projected to 2D. In a word, the object world is in 3D, we believe that building models on 3D is quite important.

{\textbf{Comparisons to Dense Correspondence Exploring.}}  We modify our 3D keypoint offset module $\mathbfcal{M_K}$ to output the corresponding 3D coordinate of each point in the object coordinate system and apply the least-squares fitting algorithm to computes the 6DoF pose. An L1 loss similar to Formula \ref{eqn:Lctr} is applied to supervise the training of the corresponding 3D coordinate. Evaluation results are shown as Ours(corr) in Tabel \ref{tab:YCB_FM}, which shows that our 3D keypoints formulation still performs quite better. We believe that regressing object coordinates is more difficult than keypoint detection. Because the model has to recognize each point of a mesh in the image and memorize its coordinate in the object coordinate system. However, detecting keypoints on objects in the camera system is easier since many keypoints are visible and the model can aggregate scene context in the vicinity of them.

\newcommand{\gwC}{16.0}
\begin{figure*}[htbp]
\centering
\includegraphics[width=\gwC cm]{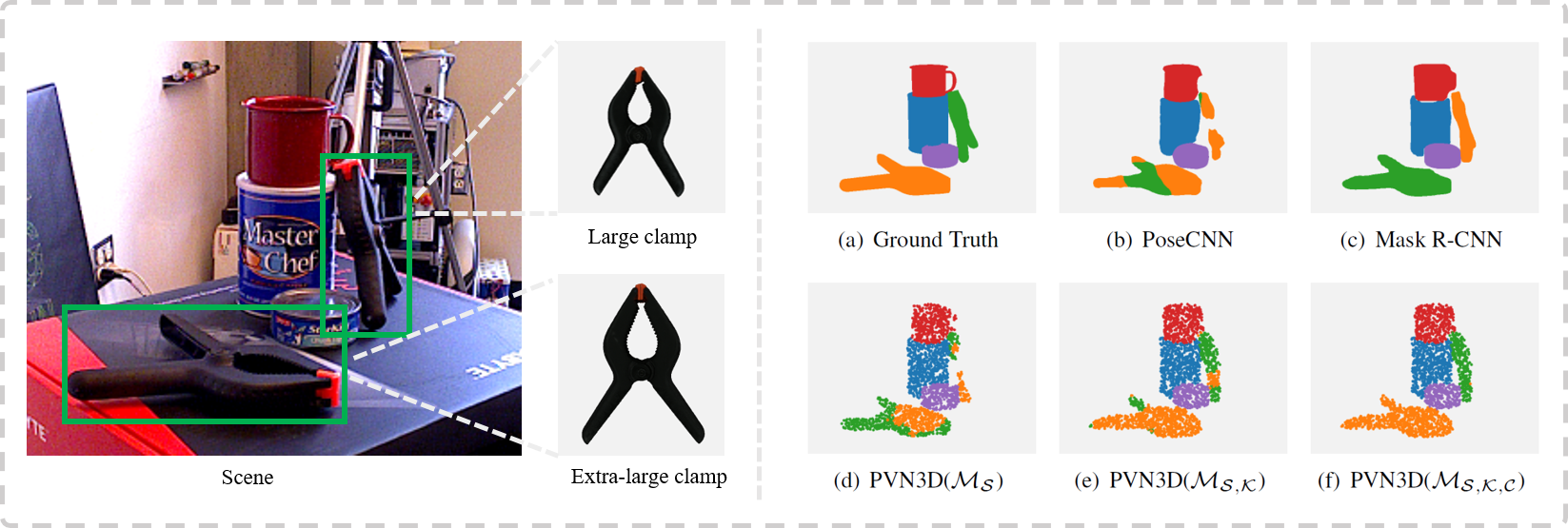}
\caption{
    \textbf{Qualitative results of semantic segmentation on the challenging YCB-Video dataset.} (a) shows the ground truth label. Different objects are labeled in different colors, with large clamp colored green and extra-large clamp colored orange. In (b)-(c), the simple baselines PoseCNN\cite{xiang2017posecnn} and Mask R-CNN \cite{maskrcnn} are confused by the two objects. In (d), our semantic segmentation module $\mathbfcal{M_S}$, trained separately, can not distinguish them well either. In (e), jointly training $\mathbfcal{M_S}$ with the keypoints offset voting module $\mathbfcal{M_K}$ performs better. In (f), with the voted center and Mean-Shift clustering algorithm, our model can distinguish them well.
}
\label{fig:segmentation}
\end{figure*}

{\textbf{Effect of 3D Keypoints Selection.}} In this part, we select 8 corners of the 3D bounding box and compares them with points selected from the FPS algorithm. Different number of keypoints generated by FPS are also taken into consideration. Table \ref{tab:KP_SL} shows that keypoints selected by the FPS algorithm on the object enable our model to perform better. That's because the bounding box corners are virtual points that are far away from points on the object. Therefore, point-based networks are difficult to aggregate scene context in the vicinity of these virtual corner points. Also, 8 keypoints selected from FPS algorithm is a good choice for our network to learn. More keypoints may better eliminate errors when recovering pose in the least-squares fitting module, but harder for the network to learn as the output space is bigger. Selecting 8 keypoints is a good trade-off.

\begin{table}[tp]
    \centering
    \fontsize{7.3}{6.5}\selectfont
    \begin{tabular}{l|C{1.0cm}|C{1.0cm}|C{1.0cm}|C{1.0cm}|C{1.0cm}}
    
        \hline
        \       &$\mathbfcal{M_K}$ +MRC  & $\mathbfcal{M_K}$ +GT    &$\mathbfcal{M_{K, S}}$ +GT & $\mathbfcal{M_{K, S, C}}$ &$\mathbfcal{M_{K, S, C}}$ +GT  \cr\hline
        ADD-S       &93.5        & 94.8	  &95.2  & 95.5  &\bf{95.7} \cr
        ADD(S)      &89.7        & 90.6	  &91.3  & 91.8  &\bf{91.9} \cr
        \hline  
    \end{tabular}
    \caption{Performance of PVN3D with different instance semantic segmentation on all objects in the YCB-Video dataset. $\mathbfcal{M_K, M_S}$ and $\mathcal{M_C}$ denote keypoint offset module, semantic segmentation and center point offset module of PVN3D respectively. +MRC and +GT denotes inference with segmentation result of Mask R-CNN and ground truth segmentation respectively. 
    }
    \label{tab:SEG_Pose}
\end{table}

\begin{table}[tp]
    \centering
    \fontsize{7.3}{6.5}\selectfont
    \begin{tabular}{L{1.3cm}|C{0.9cm}|C{0.9cm}|C{0.9cm}|C{0.9cm}|C{0.94cm}}
    \hline
                                              &PoseCNN \cite{xiang2017posecnn} & Mask R-CNN\cite{maskrcnn} & PVN3D ($\mathbfcal{M_S}$) & PVN3D ($\mathbfcal{M_{S, K}}$) & PVN3D ($\mathbfcal{M_{S, K, C}}$) \cr\hline
    large clamp          & 43.1 & 48.4       & 58.6             & 62.5           & \bf{70.2}                  \cr\hline
    extra-large clamp    & 30.4 & 36.1       & 41.5             & 50.7           & \bf{69.0}                  \cr\hline
    \end{tabular}
    \caption{Instance semantic segmentation results (mIoU(\%)) of different methods on the YCB-Video dataset. Jointly training semantic segmentation module with keypoint offset module ($\mathbfcal{M_{S,K}}$) obtains size information from the offset module and performs better, especially on large clamp and extra-large clamp. With the center voting module $\mathbfcal{M_C}$ and the Mean-Shift clustering algorithm, further improvement of performance is obtained.}
    \label{tab:SEG}
\end{table}

{\textbf{Effect of Multi-task learning.}} In this part, we discuss how the joint learning of semantic segmentation and keypoint (center) translation offset boosts the performance. In Table \ref{tab:SEG_Pose}, we explore how semantic segmentation enhances keypoint offset learning. We remove semantic segmentation and center voting modules $\mathbfcal{M_S, M_C}$, and train our keypoint voting module $\mathbfcal{M_K}$ individually. During inference time, the instance semantic segmentation predicted by Mask R-CNN \cite{maskrcnn} ($\mathbfcal{M_K}$+MRC) and the ground truth ($\mathbfcal{M_K}$+GT) are applied. Experimental results show that jointly trained with semantic segmentation ($\mathbfcal{M_{K,S}}$+GT) boosts the performance of keypoint offset voting and improves the accuracy of 6D pose estimation by 0.7\% on ADD(S) metric. We believe that the semantic module extracts global and local features to distinguish different objects. Such features also help the model to recognize which part of an object a point belongs to and improve offset prediction. 

In Table \ref{tab:SEG}, we explore how keypoint and center point offset learning improve the instance semantic segmentation result. Point mean intersection over union (mIoU) is used as evaluation metric. We report the results of the challenging large clamp and extra-large clamp in YCB-Video dataset. They look same in appearance but are different in size, as shown in Figure \ref{fig:segmentation}. We trained Mask R-CNN (ResNeXt-50-FPN) \cite{maskrcnn} with the recommended setting as a simple baseline and found it was completely confused by the two objects. With extra depth information, our semantic segmentation module (PVN3D($\mathbfcal{M_S}$)), trained individually, didn't perform well either. However, jointly trained with our keypoint offset voting module (PVN3D($\mathcal{M_{S,K}}$)), the mIoU was improved by 9.2\% on the extra-large clamp. With voted centers obtained from the center voting module $\mathbfcal{M_C}$, we can split up objects with the Mean-Shift clustering algorithm and assign points to its closest object cluster. The mIoU of the extra-large clamp is further improved by 18.3\% in this way. Some qualitative results are shown in Figure \ref{fig:segmentation}.

%% file: discussion.tex
\section{Conclusion}
We propose a novel deep 3D keypoints voting network with instance semantic segmentation for 6DoF pose estimation, which outperforms all previous approaches in several datasets by large margins. We also show that jointly training 3D keypoint with semantic segmentation can boost the performance of each other. We believe the 3D keypoint based approach is a promising direction to explore for the 6DoF pose estimation problem.  


%% file: appendix.tex
\section{Appendix}

\renewcommand{\arraystretch}{1.5}
\newcommand{\tbw}{1.4}
\begin{table*}[htbp]
    \fontsize{7.9}{7.8}\selectfont
    \begin{tabular}{l|C{\tbw cm}|C{\tbw cm}|C{\tbw cm}|C{\tbw cm}|C{\tbw cm}}
        \hline
         & PoseCNN \cite{xiang2017posecnn} & Mask R-CNN \cite{maskrcnn} & $\mathcal{M_S}$ & $\mathcal{M_{S,K}}$ & $\mathcal{M_{S,K,C}}$ \cr\hline
        002\_master\_chef\_can   & 86.9 & 87.7          & 88.0          & \textbf{88.1} & 87.7          \\
        003\_cracker\_box        & 87.5 & 85.9          & 88.0          & 88.2          & \textbf{88.3} \\
        004\_sugar\_box          & 92.0 & 91.1          & \textbf{92.3} & 91.4          & 91.5          \\
        005\_tomato\_soup\_can   & 86.4 & 87.1          & 88.5          & \textbf{88.7} & \textbf{88.7} \\
        006\_mustard\_bottle     & 93.1 & 93.1          & 92.7          & 93.2          & \textbf{93.3} \\
        007\_tuna\_fish\_can     & 89.2 & 88.8          & 89.2          & 89.5          & \textbf{89.6} \\
        008\_pudding\_box        & 73.9 & 82.5          & 86.7          & 90.7          & \textbf{90.8} \\
        009\_gelatin\_box        & 84.3 & 91.9          & \textbf{92.6} & 89.5          & 89.5          \\
        010\_potted\_meat\_can   & 83.9 & 84.9          & \textbf{88.4} & 86.4          & 86.3          \\
        011\_banana              & 90.0 & \textbf{90.2} & 89.0          & 89.0          & 89.1          \\
        019\_pitcher\_base       & 95.7 & 92.2          & 96.1          & 96.3          & \textbf{96.4} \\
        021\_bleach\_cleanser    & 88.2 & 90.2          & 87.1          & \textbf{91.8} & 91.7          \\
        024\_bowl                & 90.3 & 90.8          & \textbf{92.1} & 79.0          & 78.8          \\
        025\_mug                 & 83.0 & 80.7          & 83.8          & \textbf{84.1} & \textbf{84.1} \\
        035\_power\_drill        & 86.4 & 87.8          & 87.2          & \textbf{88.8} & \textbf{88.8} \\
        036\_wood\_block         & 79.8 & \textbf{83.2} & 82.6          & 78.5          & 78.5          \\
        037\_scissors            & 65.7 & 51.6          & 67.1          & 77.1          & \textbf{77.2} \\
        040\_large\_marker       & 69.7 & 73.4          & \textbf{76.9} & 74.5          & 74.6          \\
        051\_large\_clamp        & 43.1 & 48.4          & 58.5          & 62.5          & \textbf{70.2} \\
        052\_extra\_large\_clamp & 30.4 & 36.1          & 41.5          & 50.7          & \textbf{69.0} \\
        061\_foam\_brick         & 87.7 & 87.4          & \textbf{89.3} & 87.4          & 87.3          \cr\hline
        MEAN                     & 80.3 & 81.2          & 83.7          & 84.1          & \textbf{85.3}   \cr\hline
    \end{tabular}
    \centering
    \caption{Instance semantic segmentation results (mIoU(\%)) of different approaches.}
    \label{tab:instance_seg}
\end{table*}


\subsection{Architecture Details.}
The image embedding network in the Feature Extraction module consists of a standard ResNet-34 encoder pre-trained with ImageNet and followed by 4 up-sampling layers as a decoder. The output RGB feature is of 128 channels. The point cloud embedding network is a PointNet++ \cite{qi2017pointnet++} with Multi-scale Grouping (MSG), the output of which is also of 128 channels. In the DenseFuion block, for each sampled point, the 128 channels RGB feature is concatenated with the corresponding 128 channels point cloud feature and is fed into shared MLPs to raise the feature channels to 1024. The 128 channels RGB and point cloud feature are also fed into shared MLPs and raised to 256 channels respectively. They are then all concatenated together to form the 1792 (128*2+256*2+1024) channels RGBD features. The 3D keypoint offset voting module $\mathcal{M_K}$ consists of shared MLPs (1792-1024-512-128-$N_{kp}$*3) reducing the 1792 channels RGBD features to $N_{kp}$ 3D keypoints offset. The semantic segmentation module $\mathcal{M_S}$ consists of shared MLPs (1792-1024-512-128-$N_{cls}$) with $N_{cls}$ the number of object classes. The center voting module $\mathcal{M_C}$ also consists of shared MLPs (1792-1024-512-128-3). During training, Adam optimization algorithm is employed, with a mini-batch size of 24. We applied cyclical learning rates during training, the range of which is from 1e-5 to 1e-3.

\subsection{Implementation: Models for LineMOD dataset.}
In the LineMOD dataset, though there are multi objects in one scene, only the label of one target object in each scene is provided, making it hard to train a single model for all objects. Therefore, we follow PVNet\cite{peng2019pvnet} and trained models separately for each object. It means our semantic segmentation module $\mathcal{M_S}$ only predicts two semantic labels for each object, label of the target object and the background.

\subsection{Implementation: PVN3D+ICP}
We introduce our implementation of PVN3D+ICP as follows. During training and inference of PVN3D, we randomly sample 12288 points (pixels) from the whole scene as input. Only these points are labeled with the semantic label by our instance semantic segmentation module, which is not enough for a good performance of the ICP algorithm. Therefore, for each unlabeled point in the whole point cloud scene, we find its closest labeled point and assigned that label to it. Points with the same instance semantic labels are then selected. To eliminate the effect of noise points, a MeanShift \cite{comaniciu2002mean} clustering algorithm is also applied. The biggest cluster of points is then selected as the input of the ICP algorithm.

\subsection{More Results}

\subsubsection{Instance semantic segmentation}
We provide more results of instance semantic segmentation in Table \ref{tab:instance_seg}.

\subsubsection{Visualization of detected keypoints}
The visualization of some detected keypoints is shown in Figure \ref{fig:keypoints_vis}.

\subsubsection{Evaluation on the Occlusion LineMOD Dataset}
The Occlusion LineMOD dataset \cite{brachmann2014learning} is additionally annotated from the LineMOD datasets, objects of which are heavily occluded, making it harder for pose estimation. Evaluation results in Table \ref{tab:OCC_LM_PF} show that our PVN3D outperforms previous methods by a large margin.

\renewcommand{\arraystretch}{1.5}
\newcommand{\astrch}{1.3}
\begin{table*}[htbp]
    \fontsize{8.0}{7.8}\selectfont
    \begin{tabular}{l|C{\astrch cm}|C{\astrch cm}|C{\astrch cm}|C{\astrch cm}|C{\astrch cm}|C{\astrch cm}|C{\astrch cm}}
        \hline
                        & PoseCNN \cite{xiang2017posecnn} & Oberweger \cite{oberweger2018making} & Hu et al. \cite{hu2019segmentation}   & Pix2Pose \cite{park2019pix2pose} & PVNet \cite{peng2019pvnet} & DPOD \cite{DPOD} & PVN3D         \cr \hline
        ape             & 9.6     & 12.1      & 17.6 & 22.0     & 15.8  & -    & \textbf{33.9} \\
        can             & 45.2    & 39.9      & 53.9 & 44.7     & 63.3  & -    & \textbf{88.6} \\
        cat             & 0.9     & 8.2       & 3.3  & 22.7     & 16.7  & -    & \textbf{39.1} \\
        driller         & 41.4    & 45.2      & 62.4 & 44.7     & 65.7  & -    & \textbf{78.4} \\
        duck            & 19.6    & 17.2      & 19.2 & 15.0     & 25.2  & -    & \textbf{41.9} \\
        \textbf{eggbox} & 22.0    & 22.1      & 25.9 & 25.2     & 50.2  & -    & \textbf{80.9} \\
        \textbf{glue}   & 38.5    & 35.8      & 39.6 & 32.4     & 49.6  & -    & \textbf{68.1} \\
        holepuncher     & 22.1    & 36.0      & 21.3 & 49.5     & 39.7  & -    & \textbf{74.7} \cr\hline
        average         & 24.9    & 27.0      & 27.0 & 32.0     & 40.8  & 47.3 & \textbf{63.2} \cr\hline
    \end{tabular}
    \centering
    \caption{Quantitative evaluation of 6D Pose on ADD(S) \cite{hinterstoisser2012model} metric on the Occlusion LineMOD dataset. Objects with bold name are symmetric.}
    \label{tab:OCC_LM_PF}
\end{table*}

\subsubsection{Inference time}
It takes 0.17 seconds for network forward propagation and 0.02 seconds for pose estimation of each object during inference. The overall runtime is 5 FPS on the LineMOD dataset.

\newcommand{\gwCk}{15.5}
\begin{figure*}[htbp]
    \centering
    \includegraphics[width=\gwCk cm]{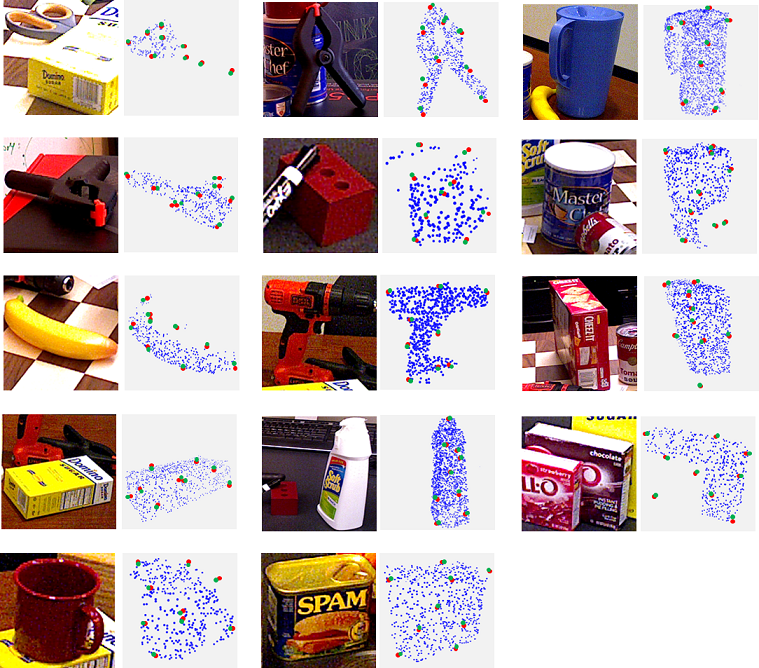}
    \caption{
        \textbf{Visualization of 3D keypoints in YCB-Video dataset.} Blue points are sampled point clouds from the scene. Red points are the ground truth 3D keypoints and green points are predicted 3D keypoints.
    }
    \label{fig:keypoints_vis}
\end{figure*}